\documentclass[conference,a4paper]{IEEEtran}
\IEEEoverridecommandlockouts

\usepackage{times}
\usepackage{epsfig}
\usepackage{graphicx}
\usepackage{amsmath}
\usepackage{amssymb}
\usepackage{bm}
\usepackage{xcolor}
\usepackage{caption}
\usepackage{subcaption}
\usepackage{booktabs}
\usepackage{multirow}
\usepackage{url}

\makeatletter
\def\ps@IEEEtitlepagestyle{%
  \def\@oddfoot{\mycopyrightnotice}%
  \def\@evenfoot{}%
}
\def\mycopyrightnotice{%
  \parbox{\textwidth}{\centering\footnotesize
  Copyright 2025 IEEE. Published in the Digital Image Computing: Techniques and Applications, 2025 (DICTA 2025), 3-5 December 2025 in Adelaide, South Australia, Australia. Personal use of this material is permitted. However, permission to reprint/republish this material for advertising or promotional purposes or for creating new collective works for resale or redistribution to servers or lists, or to reuse any copyrighted component of this work in other works, must be obtained from the IEEE. Contact: Manager, Copyrights and Permissions / IEEE Service Center / 445 Hoes Lane / P.O. Box 1331 / Piscataway, NJ 08855-1331, USA. Telephone: + Intl. 908-562-3966.}
}
\makeatother

\def\overview{
\begin{figure}[t]
\centering
\includegraphics[width=0.95\linewidth]{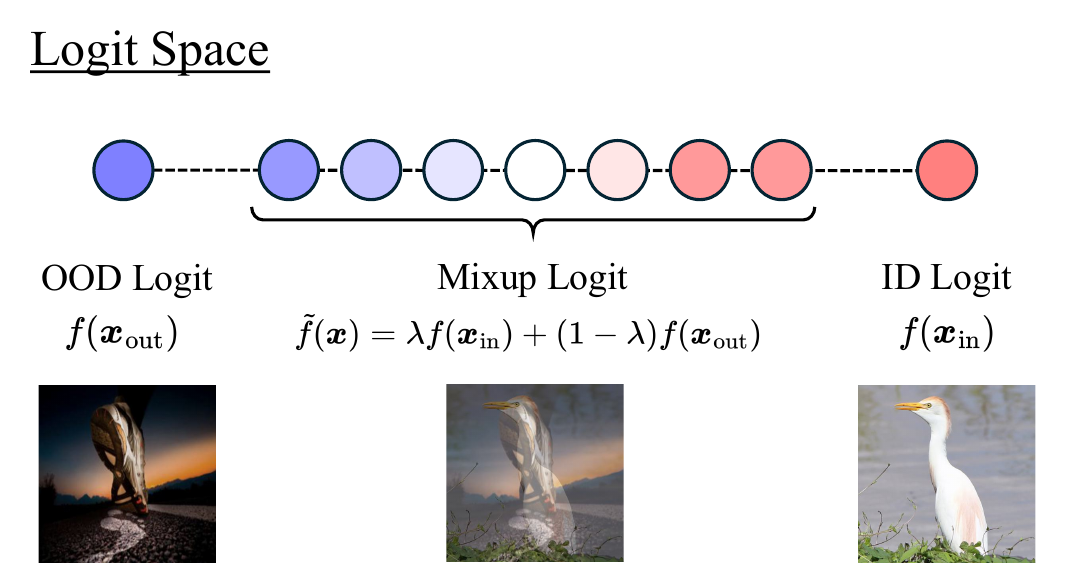}
\caption{Overview of Logit MixOE. For out-of-distribution detection, the proposed method not only mixes up the in-distribution data $\bm{x}_{\mathrm{in}}$ and out-of-distribution data $\bm{x}_{\mathrm{out}}$ in the input space, but also regularises them by mixup in the logit space.}
\label{fig:overview}
\end{figure}
}

\def\overviewlogitmix{
\begin{figure*}[t]
\centering
\includegraphics[width=0.45\linewidth]{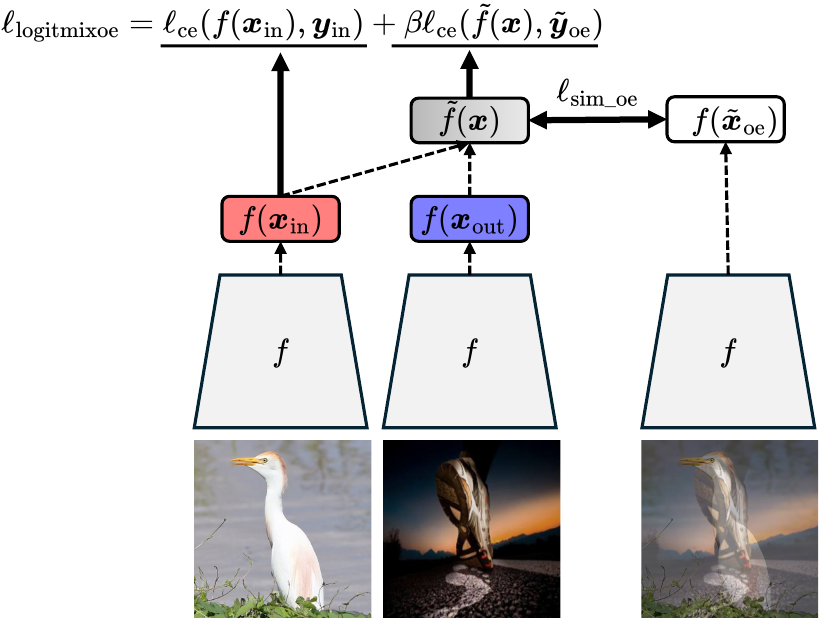}
\caption{Overall view of Logit MixOE. The proposed method learns from the out-of-distribution loss $\ell_\mathrm{logitmixoe}$ due to Outlier Exposure for mixed Logit in Logit space and the consistency loss $\ell_\mathrm{sim\_oe}$ with Logit for mixed data in input space. Consistency loss with Logit of mixed data in the input space $\ell_\mathrm{sim\_oe}$ is learned.}
\label{fig:overview_logitmixoe}
\end{figure*}
}

\newcommand{\colwidthlogit}{0.19\textwidth}
\def\figlogithist{
\begin{figure*}[t]
\centering
\begin{minipage}[b]{\colwidthlogit}{
\centering
\includegraphics[width=1.0\linewidth]{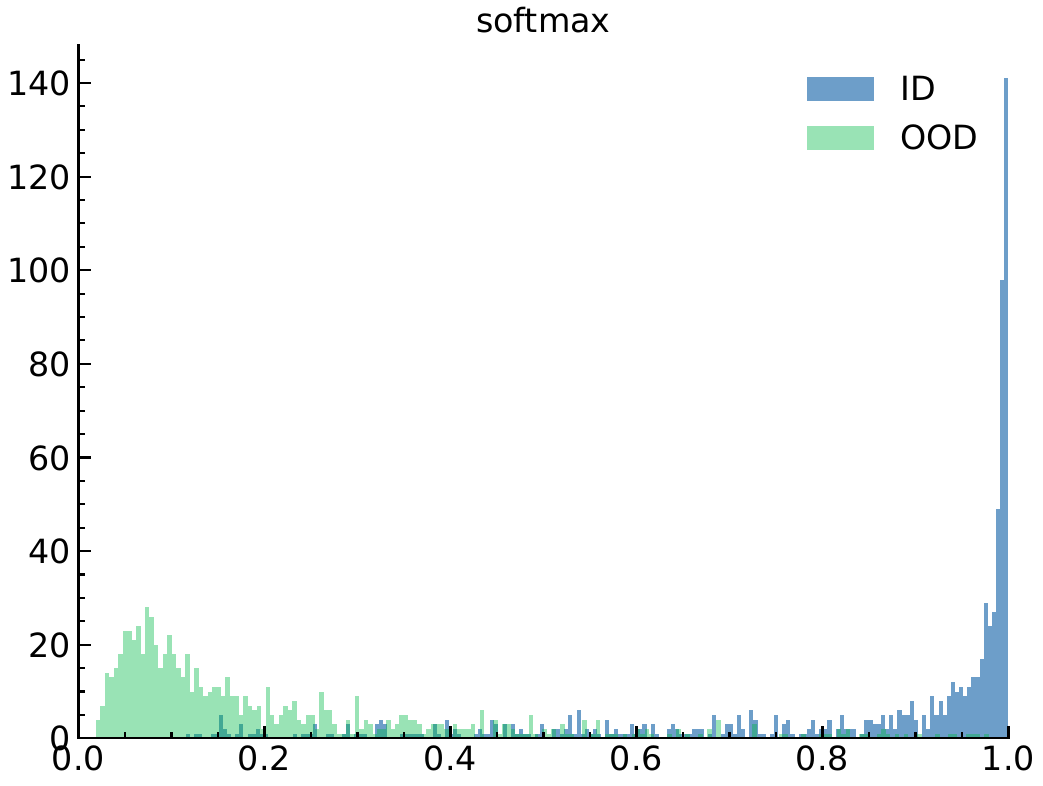}
\subcaption*{\footnotesize Pre-trained model}
}
\end{minipage}
\begin{minipage}[b]{\colwidthlogit}{
\centering
\includegraphics[width=1.0\linewidth]{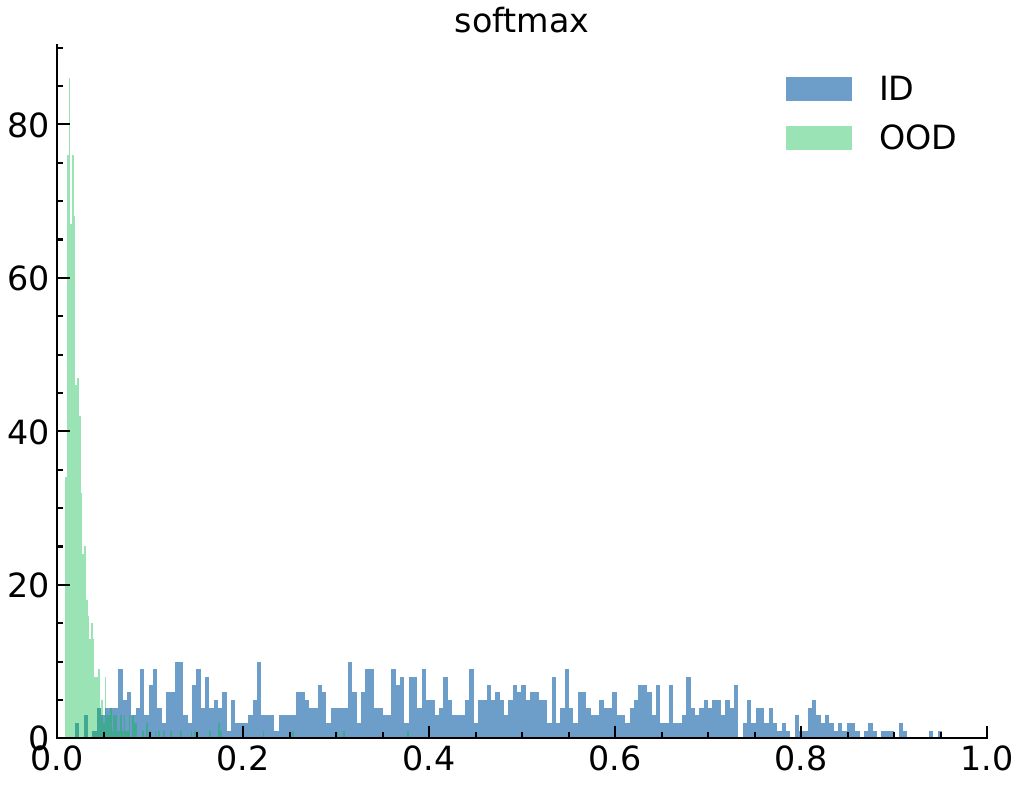}
\subcaption*{\footnotesize MixOE}
}
\end{minipage}
\begin{minipage}[b]{\colwidthlogit}{
\centering
\includegraphics[width=1.0\linewidth]{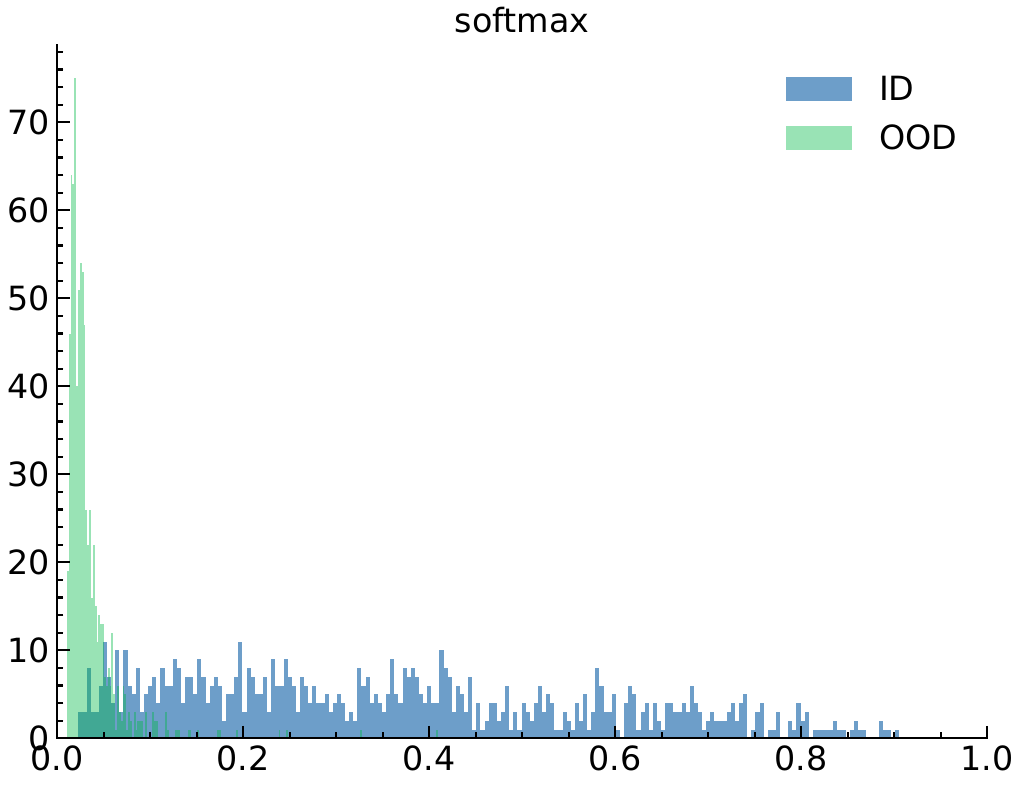}
\subcaption*{\footnotesize Logit MixOE}
}
\end{minipage}
\begin{minipage}[b]{\colwidthlogit}{
\centering
\includegraphics[width=1.0\linewidth]{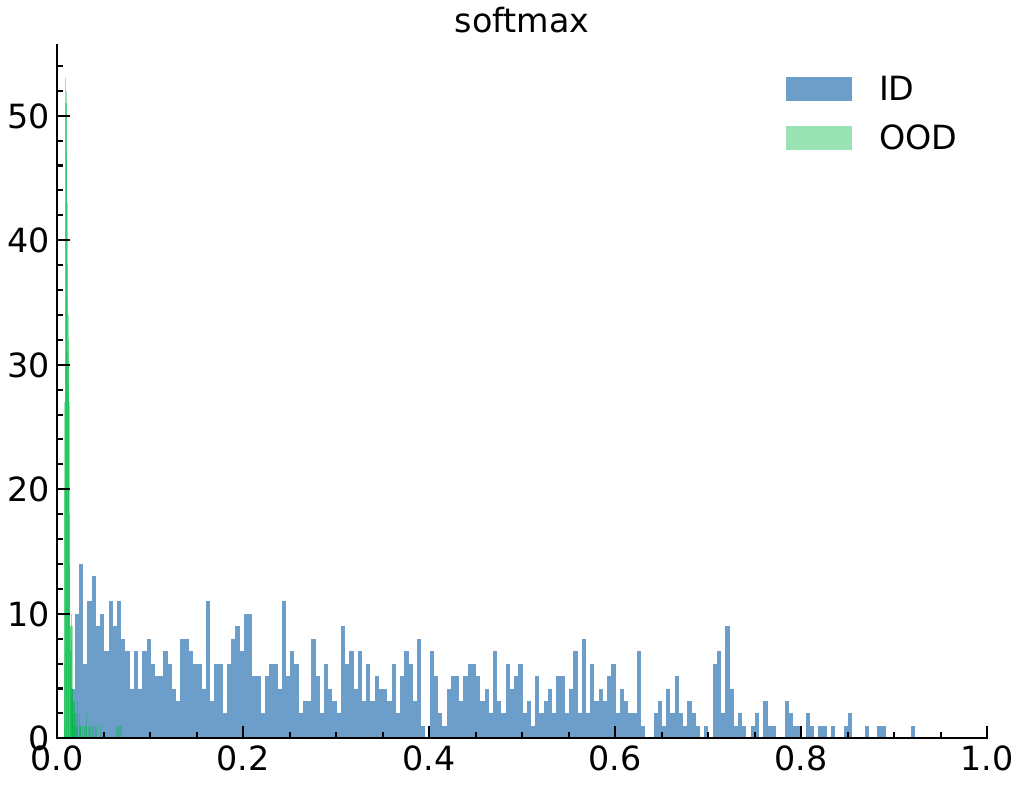}
\subcaption*{\footnotesize MixOE w/ $\ell_\mathrm{sim\_oe}$}
}
\end{minipage}
\begin{minipage}[b]{\colwidthlogit}{
\centering
\includegraphics[width=1.0\linewidth]{figures/logit_mix_oe_l_sim/hist.pdf}
\subcaption*{\footnotesize LogitMixOE w/ $\ell_\mathrm{sim\_oe}$}
}
\end{minipage}
\caption{Histogram visualisation of the L2 norm in Logit}
\label{fig:logithist}
\end{figure*}
}

\def\figlogitpca{
\begin{figure*}[t]
\centering
\begin{minipage}[b]{\colwidthlogit}{
\centering
\includegraphics[width=1.0\linewidth]{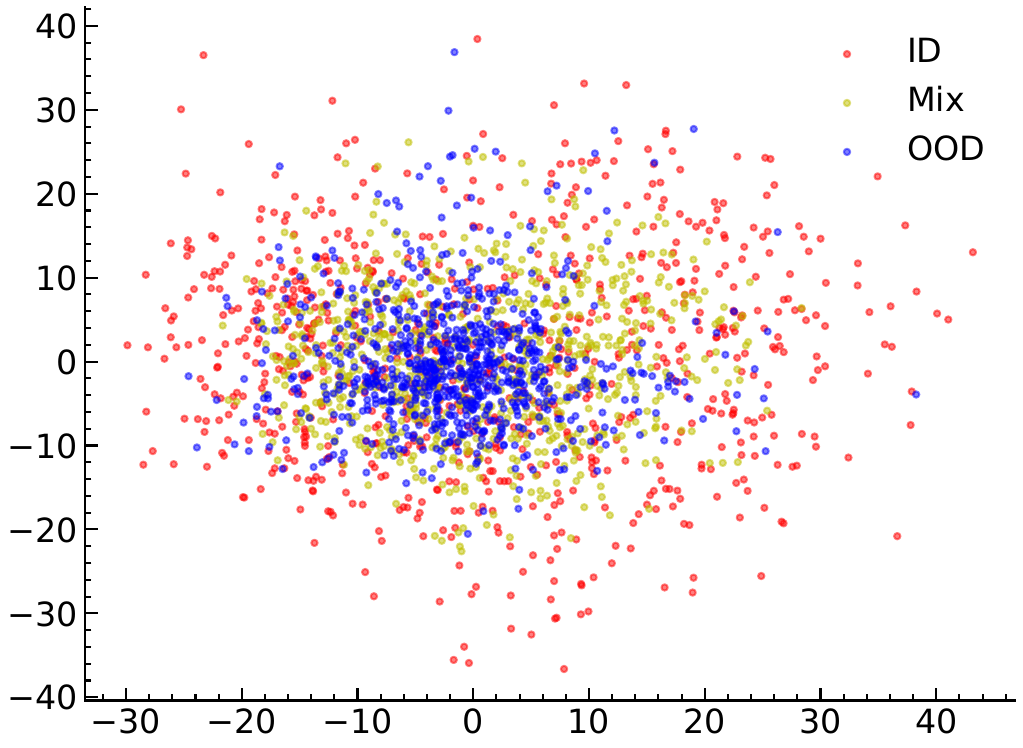}
\subcaption*{\footnotesize Pre-trained model}
}
\end{minipage}
\begin{minipage}[b]{\colwidthlogit}{
\centering
\includegraphics[width=1.0\linewidth]{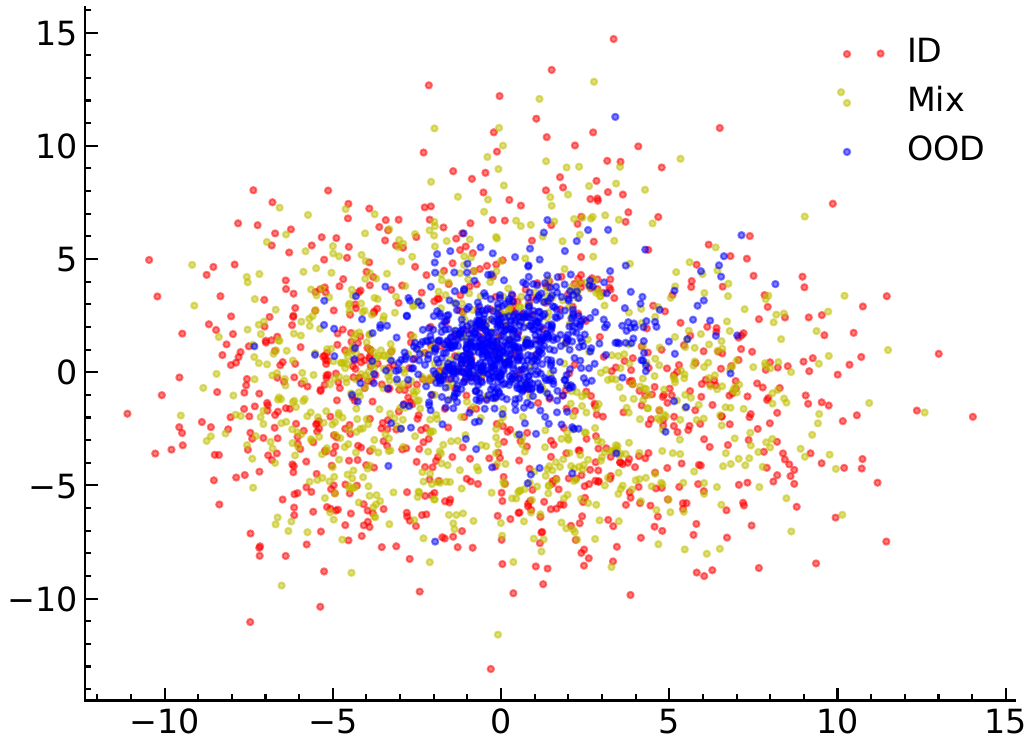}
\subcaption*{\footnotesize MixOE}
}
\end{minipage}
\begin{minipage}[b]{\colwidthlogit}{
\centering
\includegraphics[width=1.0\linewidth]{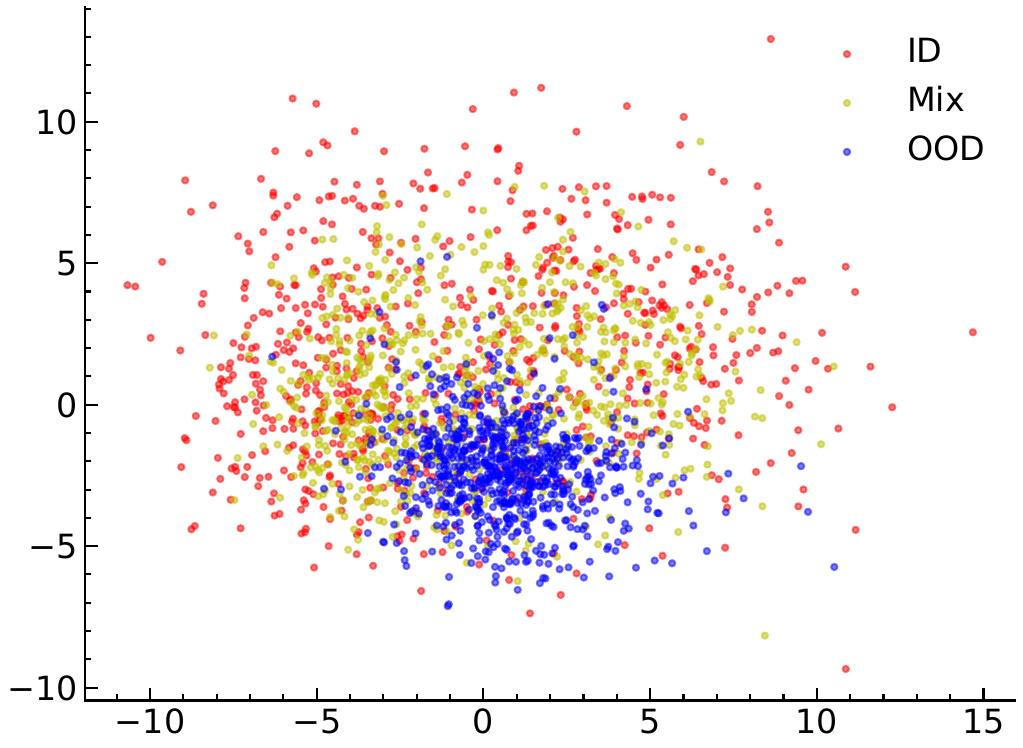}
\subcaption*{\footnotesize Logit MixOE}
}
\end{minipage}
\begin{minipage}[b]{\colwidthlogit}{
\centering
\includegraphics[width=1.0\linewidth]{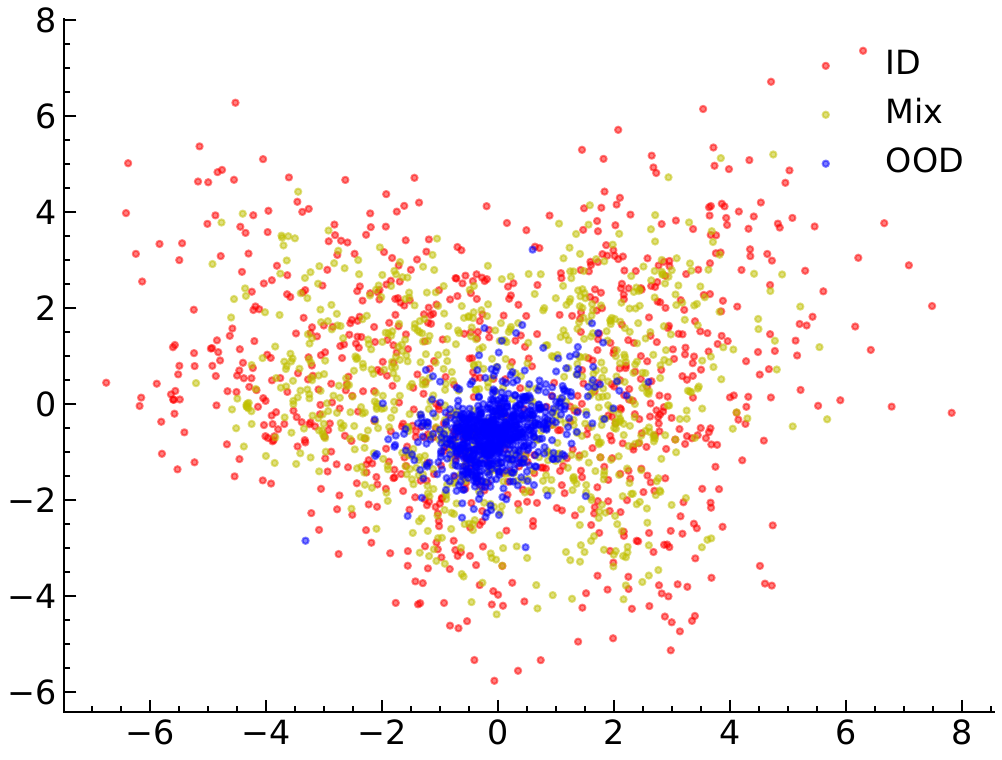}
\subcaption*{\footnotesize MixOE w/ $\ell_\mathrm{sim\_oe}$}
}
\end{minipage}
\begin{minipage}[b]{\colwidthlogit}{
\centering
\includegraphics[width=1.0\linewidth]{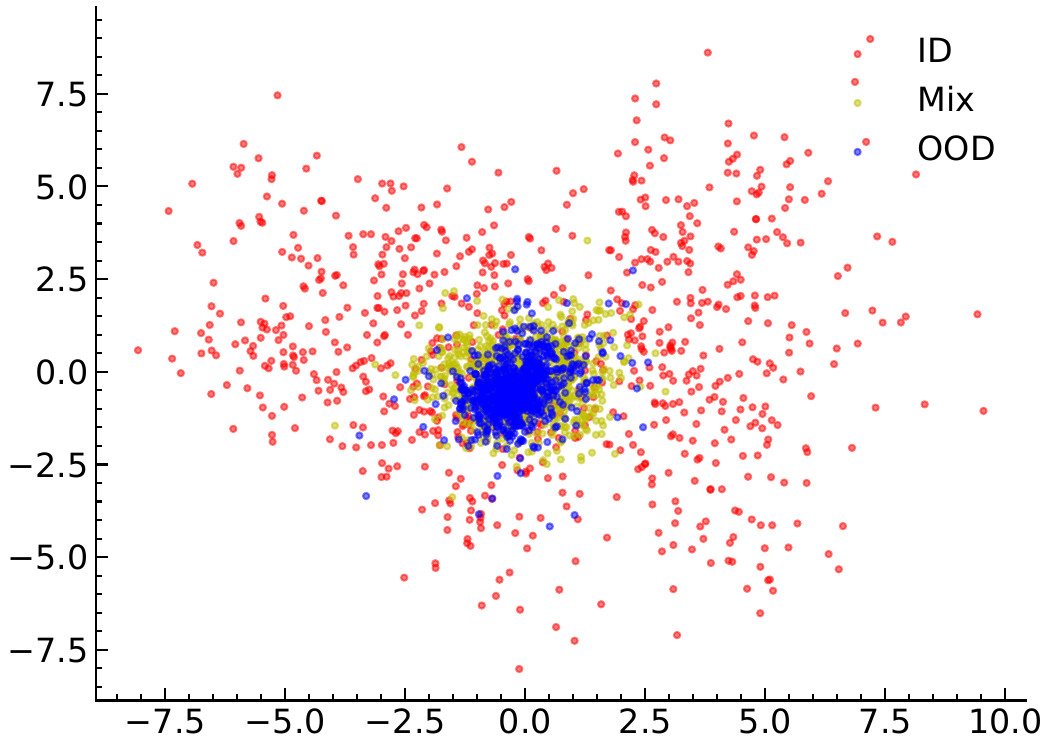}
\subcaption*{\footnotesize LogitMixOE w/ $\ell_\mathrm{sim\_oe}$}
}
\end{minipage}
\caption{Visualisation of Logit distribution by PCA}
\label{fig:logitpca}
\end{figure*}
}

\newcommand{\colwidthsample}{0.23\textwidth}

\def\figlogitlogitmixoe{
\begin{figure*}[t!]
\centering
\begin{minipage}[b]{\colwidthsample}{
\centering
\includegraphics[width=1.0\linewidth]{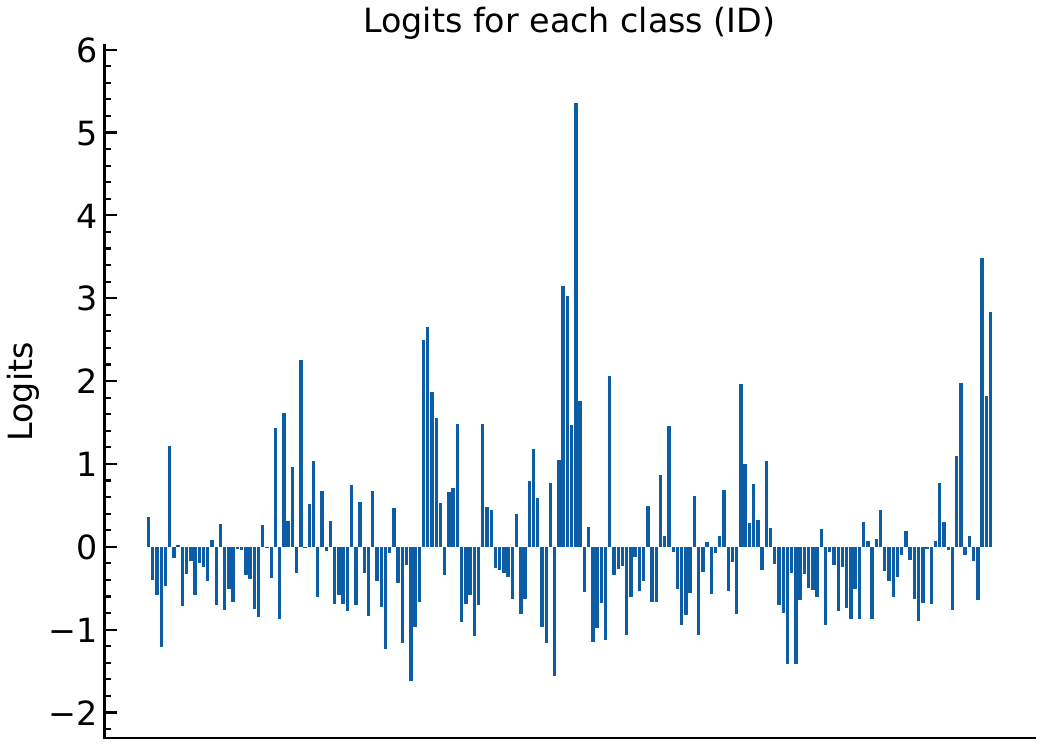}
\subcaption*{\footnotesize In-Distribution data}
}
\end{minipage}
\begin{minipage}[b]{\colwidthsample}{
\centering
\includegraphics[width=1.0\linewidth]{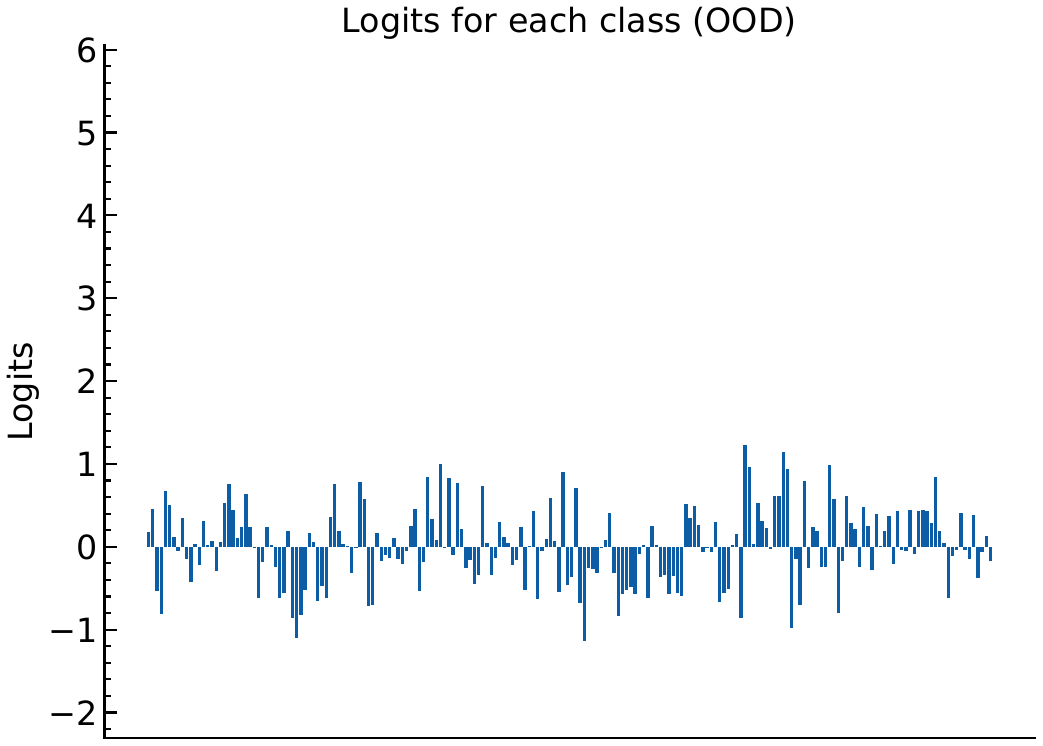}
\subcaption*{\footnotesize Out-of-Distribution data}
}
\end{minipage}
\begin{minipage}[b]{\colwidthsample}{
\centering
\includegraphics[width=1.0\linewidth]{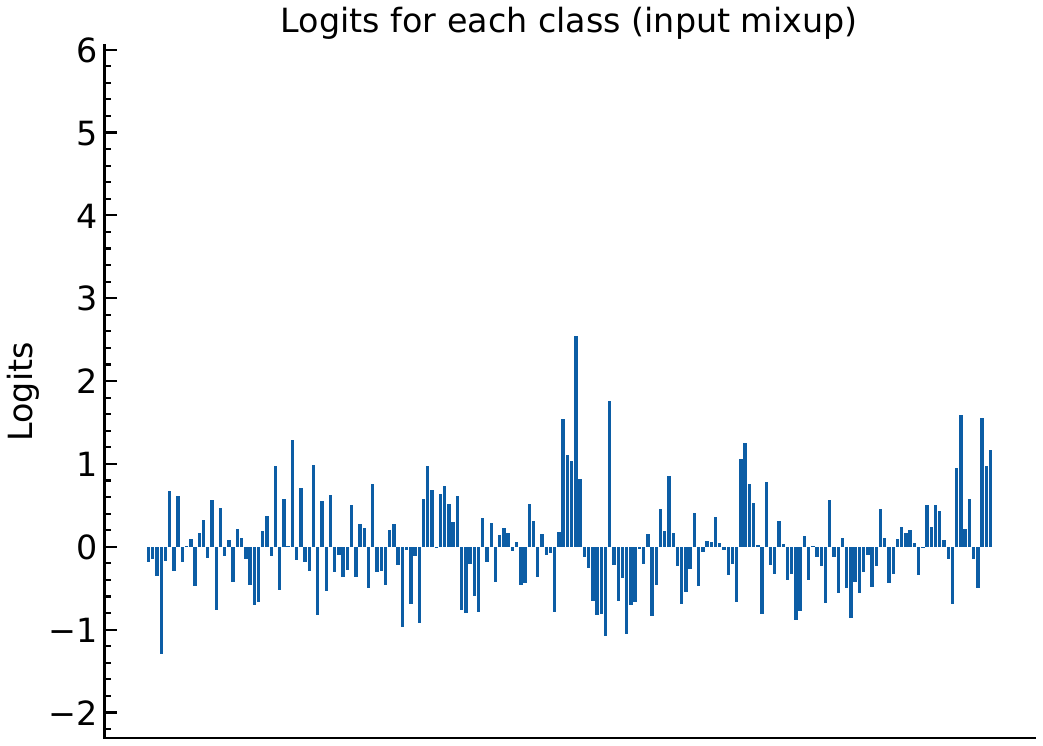}
\subcaption*{\footnotesize Mixup with input images}
}
\end{minipage}
\begin{minipage}[b]{\colwidthsample}{
\centering
\includegraphics[width=1.0\linewidth]{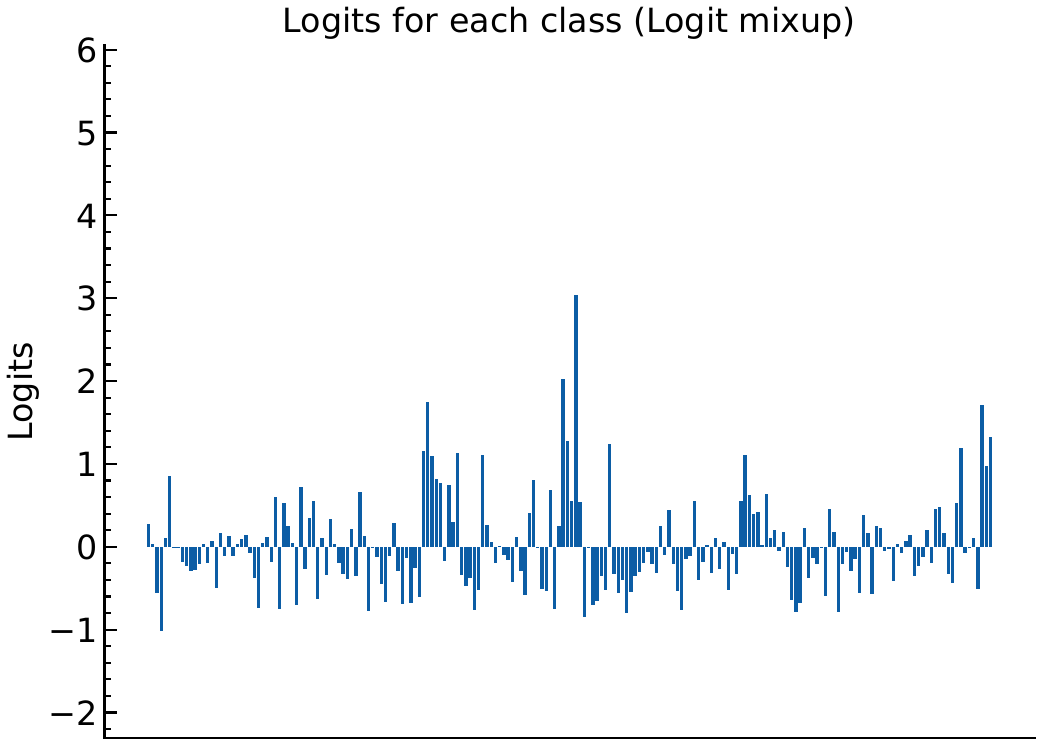}
\subcaption*{\footnotesize Mixup in Logit}
}
\end{minipage}
\caption{Visualisation of the distribution of Logit in Logit MixOE.}
\label{fig:logit_logitmixoe}
\end{figure*}
}

\def\figlogitlogitmixoelsim{
\begin{figure*}[t!]
\centering
\begin{minipage}[b]{\colwidthsample}{
\centering
\includegraphics[width=1.0\linewidth]{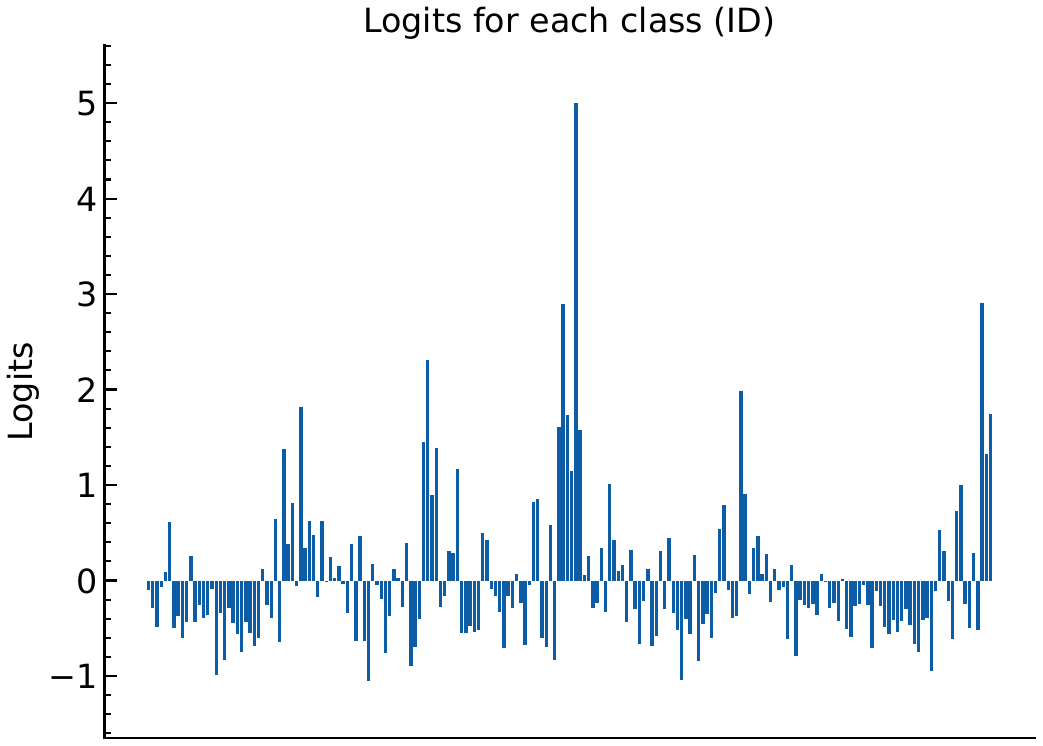}
\subcaption*{\footnotesize In-Distribution data}
}
\end{minipage}
\begin{minipage}[b]{\colwidthsample}{
\centering
\includegraphics[width=1.0\linewidth]{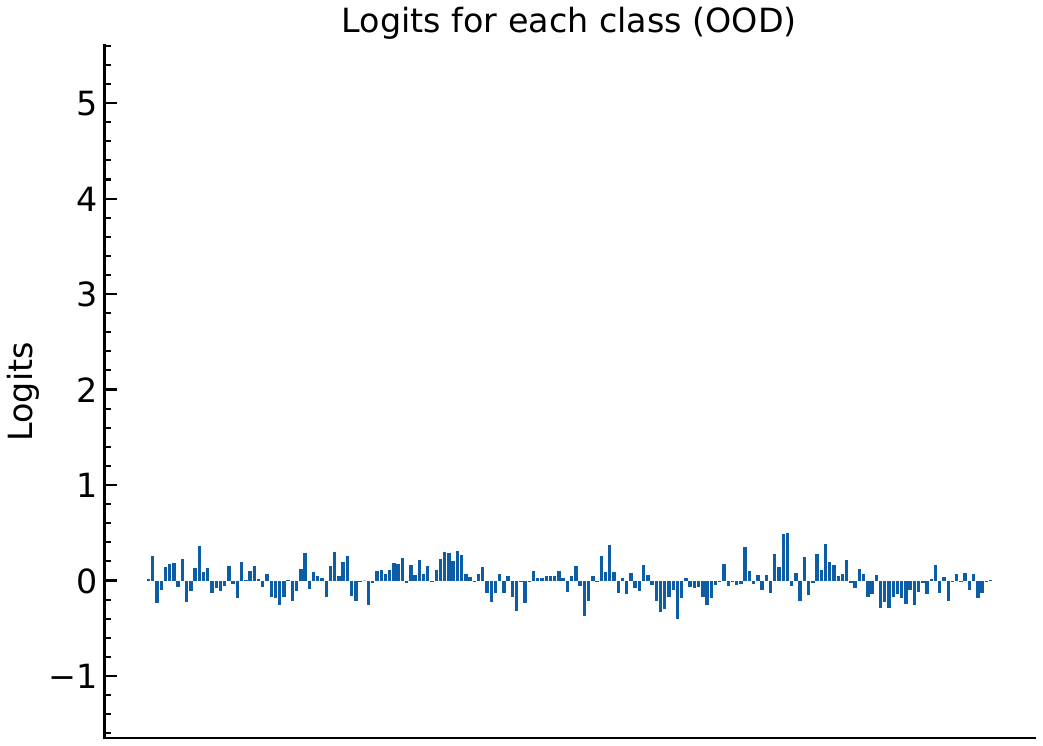}
\subcaption*{\footnotesize Out-of-Distribution data}
}
\end{minipage}
\begin{minipage}[b]{\colwidthsample}{
\centering
\includegraphics[width=1.0\linewidth]{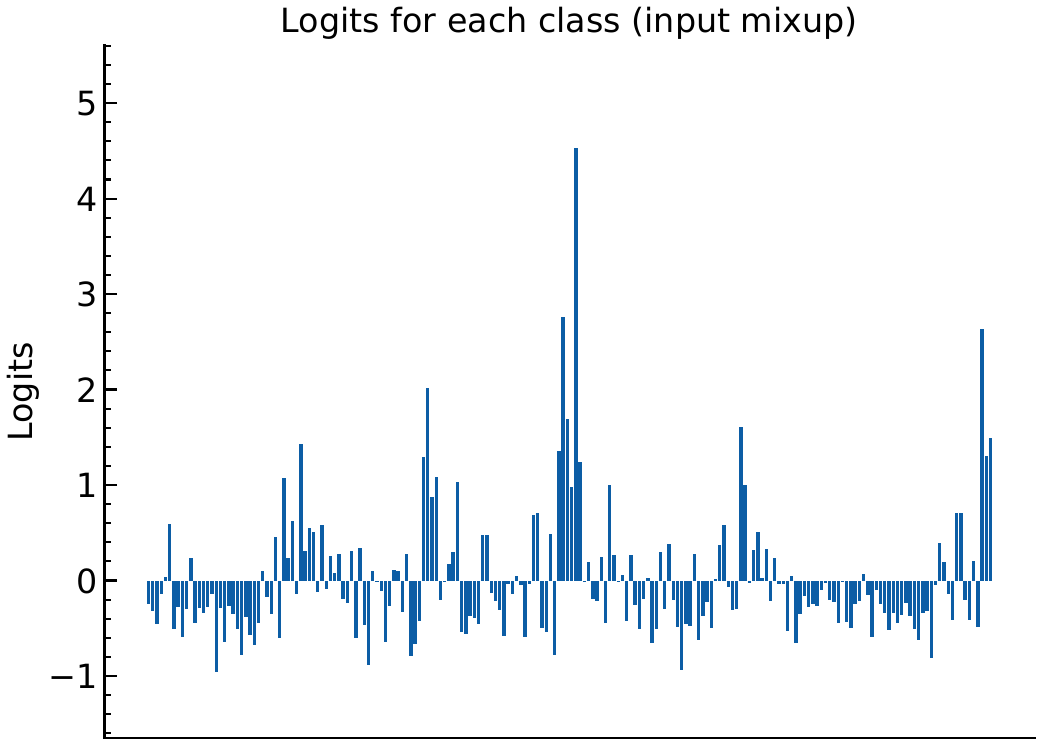}
\subcaption*{\footnotesize Mixup with input images}
}
\end{minipage}
\begin{minipage}[b]{\colwidthsample}{
\centering
\includegraphics[width=1.0\linewidth]{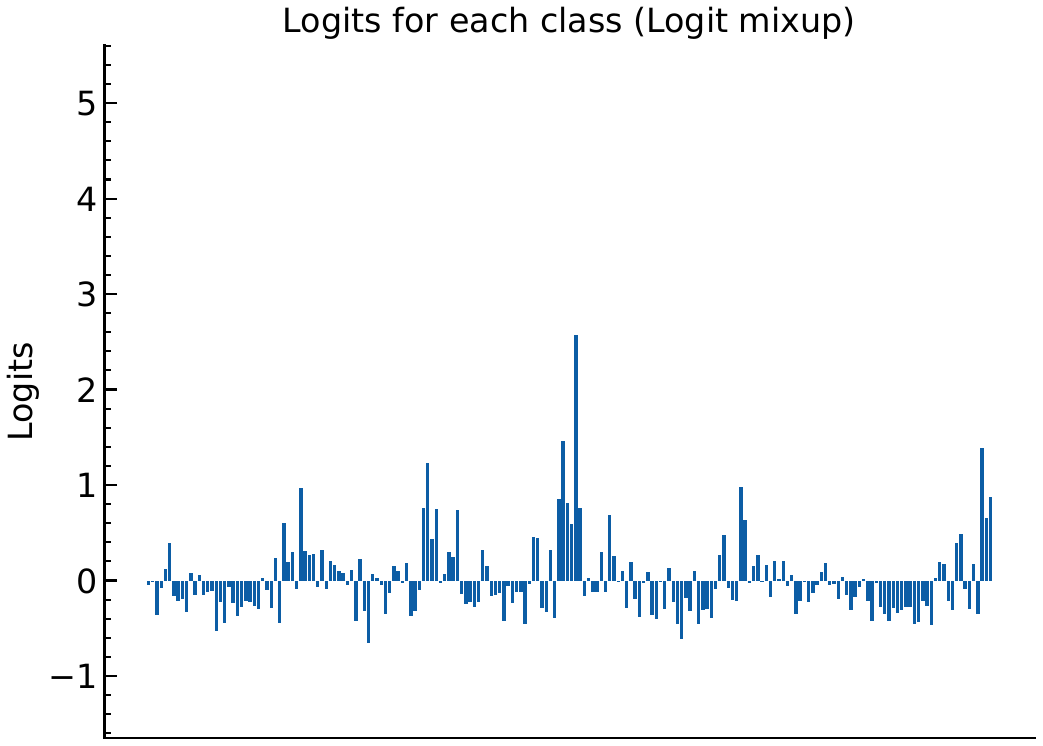}
\subcaption*{\footnotesize Mixup in Logit}
}
\end{minipage}
\caption{Visualization of the distribution of Logit in Logit MixOE w/ $\ell_\mathrm{sim\_oe}$}
\label{fig:logit_logitmixoelsim}
\end{figure*}
}

\def\figbeta{
\begin{figure}[t]
\centering
\includegraphics[width=0.75\linewidth]{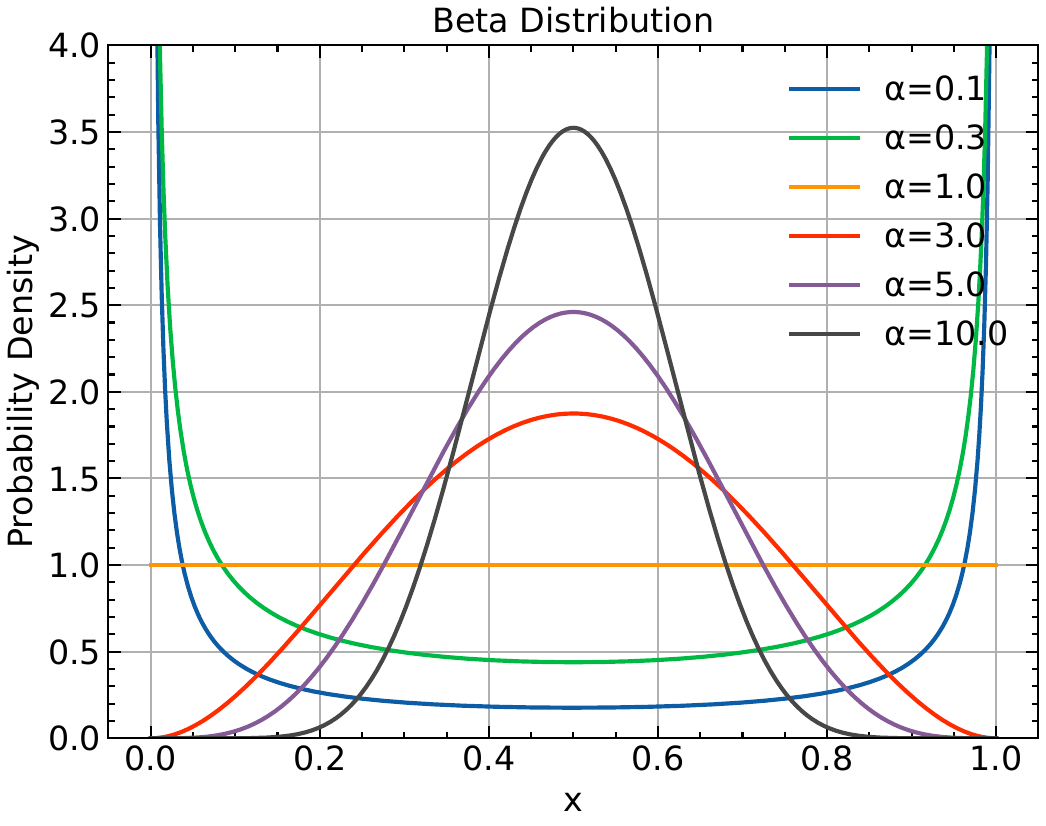}
\caption{Relationship between the Beta distribution and the hyperparameter $\alpha$.}
\label{fig:beta}
\end{figure}
}
\def\result{
\begin{table*}[t!]
\centering
\caption{Comparative results of classification performance and out-of-distribution detection performance}
\resizebox{\textwidth}{!}{
\begin{tabular}{lcccccccccc}
\toprule[0.4mm]
&  & \multicolumn{9}{c}{\textbf{ID Dataset}}\\
& & \multicolumn{3}{c}{\texttt{Bird}} & \multicolumn{3}{c}{\texttt{Aircraft}} & \multicolumn{3}{c}{\texttt{Butterfly}} \\
\textbf{Method} & $\ell_\mathrm{sim\_oe}$ & \multicolumn{1}{l}{Accuracy$\uparrow$}& \multicolumn{1}{l}{AUROC$\uparrow$} & \multicolumn{1}{l}{FPR$\downarrow$}& \multicolumn{1}{l}{Accuracy$\uparrow$} & \multicolumn{1}{l}{AUROC$\uparrow$} & \multicolumn{1}{l}{FPR$\downarrow$}& \multicolumn{1}{l}{Accuracy$\uparrow$} & \multicolumn{1}{l}{AUROC$\uparrow$} & \multicolumn{1}{l}{FPR$\downarrow$}\\ \midrule
&  & \multicolumn{9}{c}{$\alpha=$0.1} \\
\multirow{2}{*}{MixOE} & &81.12 & 0.9922 & 0.0338 & 80.38 & 0.9902 & 0.0521 & 87.11 & 0.9931 & 0.0234 \\
& \checkmark &83.25 & \textbf{0.9968} & \textbf{0.0112} & 88.54 & 0.9970 & 0.0069 & \textbf{89.06} & \textbf{0.9962} & \textbf{0.0117} \\
\multirow{2}{*}{Logit MixOE} & & 81.63 & 0.9757 & 0.1175 & 74.48 & 0.9585 & 0.2535 & 86.33 & 0.9911 & 0.0208 \\
&\checkmark & \textbf{84.63} & 0.9960 & 0.0138 & \textbf{88.72} & \textbf{0.9985} & \textbf{0.0035} & 88.15 & 0.9953 & 0.0143 \\ \hline
&  & \multicolumn{9}{c}{$\alpha=$0.3} \\
\multirow{2}{*}{MixOE} & &83.88 & 0.9955 & 0.0163 & 87.32 & 0.9950 & 0.0104 & 88.02 & 0.9951 & 0.0130 \\
& \checkmark & 83.63 & \textbf{0.9979} & \textbf{0.0100} & \textbf{89.76} & 0.9978 & 0.0104 & 87.63 & 0.9958 & 0.0143 \\
\multirow{2}{*}{Logit MixOE} & & \textbf{84.38} & 0.9803 & 0.1125 & 78.65 & 0.9583 & 0.2361 & 84.90 & 0.9867 & 0.0534 \\
& \checkmark & 82.88 & 0.9967 & 0.0175 & 88.72 & \textbf{0.9996} & \textbf{0.0017} & \textbf{88.80} & \textbf{0.9961} & \textbf{0.0104} \\ \hline
&  & \multicolumn{9}{c}{$\alpha=$1.0} \\
\multirow{2}{*}{MixOE} & &\textbf{84.88} & 0.9932 & 0.0350 & 88.72 & 0.9945 & 0.0122 & 88.54 & \textbf{0.9954} & 0.0143 \\
& \checkmark & 79.63 & 0.9982 & \textbf{0.0037} & \textbf{89.24} & 0.9994 & 0.0052 & 87.63 & 0.9943 & 0.0156 \\
\multirow{2}{*}{Logit MixOE} & & 82.75 & 0.9757 & 0.1275 & 85.76 & 0.9847 & 0.0590 & \textbf{89.06} & 0.9920 & 0.0352 \\
& \checkmark & 81.50 & \textbf{0.9984} & 0.0083 & \textbf{89.24} & \textbf{0.9998} & \textbf{0.0000} & 87.63 & 0.9953 & \textbf{0.0104} \\ \hline
&  & \multicolumn{9}{c}{$\alpha=$3.0} \\
\multirow{2}{*}{MixOE} & &\textbf{85.63} & 0.9917 & 0.0362 & \textbf{89.06} & 0.9925 & 0.0156 & \textbf{88.67 }& 0.9937 & 0.0299 \\
& \checkmark & 79.25 & \textbf{0.9984} & 0.0050 & 87.33 & 0.9995 & 0.0017 & 86.85 & 0.9944 & 0.0156 \\
\multirow{2}{*}{Logit MixOE} & & 82.88 & 0.9760 & 0.1150 & 83.51 & 0.9750 & 0.1389 & 88.28 & 0.9902 & 0.0417 \\
& \checkmark & 80.75 & \textbf{0.9984} & \textbf{0.0037} & 88.02 & \textbf{0.9999} & \textbf{0.0000} & 88.02 & \textbf{0.9949} & \textbf{0.0104} \\ \hline
&  & \multicolumn{9}{c}{$\alpha=$5.0} \\
\multirow{2}{*}{MixOE} & &\textbf{85.38} & 0.9893 & 0.0525 & \textbf{88.72} & 0.9914 & 0.0295 & \textbf{89.45} & 0.9935 & 0.0326 \\
& \checkmark & 77.88 & \textbf{0.9981} & \textbf{0.0025} & 88.37 & 0.9996 & 0.0017 & 87.37 & 0.9948 & 0.0130 \\
\multirow{2}{*}{Logit MixOE} & & 82.63 & 0.9783 & 0.1237 & 86.46 & 0.9796 & 0.0972 & 87.11 & 0.9880 & 0.0534 \\
& \checkmark & 80.75 & 0.9979 & 0.0050 & \textbf{88.72} & \textbf{0.9999} & \textbf{0.0000} & 86.85 & \textbf{0.9950} & \textbf{0.0065} \\ \hline
&  & \multicolumn{9}{c}{$\alpha=$10.0} \\
\multirow{2}{*}{MixOE} & &\textbf{84.88} & 0.9883 & 0.0812 & \textbf{88.19 }& 0.9897 & 0.0330 & \textbf{89.06} & 0.9914 & 0.0430 \\
& \checkmark & 77.88 & 0.9978 & \textbf{0.0025} & 88.02 & 0.9997 & 0.0017 & 86.85 & 0.9946 & 0.0130 \\
\multirow{2}{*}{Logit MixOE} & & 83.63 & 0.9782 & 0.1375 & 84.38 & 0.9747 & 0.1545 & 86.72 & 0.9861 & 0.0625 \\
& \checkmark & 79.63 & \textbf{0.9984} & \textbf{0.0025} & 88.02 & \textbf{0.9999} & \textbf{0.0000} & 88.28 & \textbf{0.9947} & \textbf{0.0091} \\ 
\bottomrule[0.4mm]
\end{tabular}}
\label{tb:result}
\end{table*}
}

\usepackage[pagebackref=true,breaklinks=true,colorlinks,bookmarks=false]{hyperref}

\begin{document}

\title{Logit Mixture Outlier Exposure \\
for Fine-grained Out-of-Distribution Detection}


\author{Akito Shinohara\\
Hiroshima University\\
Hiroshima, Japan\\
{\tt\small m252486@hiroshima-u.ac.jp}
\and
Kohei Fukuda\\
Hiroshima University\\
Hiroshima, Japan\\
{\tt\small kohei.fukuda41@gmail.com}
\and
Hiroaki Aizawa\\
Hiroshima University\\
Hiroshima, Japan\\
{\tt\small hiroaki-aizawa@hiroshima-u.ac.jp}
}

\maketitle

\begin{abstract}
   The ability to detect out-of-distribution data is essential not only for ensuring robustness against unknown or unexpected input data but also for improving the generalization performance of the model. Among various out-of-distribution detection methods, Outlier Exposure and Mixture Outlier Exposure are promising approaches that enhance out-of-distribution detection performance by exposing the outlier data during training. However, even with these sophisticated techniques, it remains challenging for models to learn the relationships between classes effectively and to distinguish data sampling from in-distribution and out-of-distribution clearly. Therefore, we focus on the logit space, where the properties between class-wise distributions are distinctly separated from those in the input or feature spaces. Specifically, we propose a linear interpolation technique in the logit space that mixes in-distribution and out-of-distribution data to facilitate smoothing logits between classes and improve the out-of-distribution detection performance, particularly for out-of-distribution data that lie close to the in-distribution data. Additionally, we enforce consistency between the logits obtained through mixing in the logit space and those generated via mixing in the input space. Our experiments demonstrate that our logit-space mixing technique reduces the abrupt fluctuations in the model outputs near the decision boundaries, resulting in smoother and more reliable separation between in-distribution and out-of-distribution data. Furthermore, we evaluate the effectiveness of the proposed method on a fine-grained out-of-distribution detection task.
\end{abstract}

\section{Introduction}
\label{sec:outline}
The reliability of prediction is essential in safety-critical domains such as medical imaging, autonomous driving, and cybersecurity. In these settings, prediction models are typically trained on a specific dataset, known as in-distribution (ID) data, which represents the types of inputs the model is expected to encounter. However, in real-world scenarios, models are often exposed to inputs that differ significantly from this training distribution. Such inputs, referred to as out-of-distribution (OOD) data, come from unknown or previously unseen sources. The ability to accurately detect these OOD samples is crucial for ensuring trustworthy model behavior. The failure to identify such anomalous inputs can lead to serious safety risks and potential system failures. Furthermore, the capability to detect OOD samples is closely related to the generalization ability of the model, as it plays a crucial role in building more robust and practically reliable predictions.

\overview

To achieve OOD detection, various approaches have been proposed~\cite{MSP,ODIN,mahalanobis-based,energy-based,fukuda2024taylor}. In particular, Outlier Exposure~(OE)~\cite{OE} significantly enhanced OOD detection performance by exposing auxiliary OOD samples during training. However, OE remains challenging to clearly separate the boundaries between ID and OOD samples, especially in high-dimensional and complex feature spaces. To address this problem, regularization, such as data augmentation~\cite{mixup,cutmix}, dropout~\cite{Dropout}, early stopping~\cite{early_stopping}, and weight decay~\cite{weight_decay}, has attracted attention as a promising approach for not only image classification but also OOD detection~\cite{MixOE}. It helps prevent models from overfitting and enhances their generalization capabilities. 

Mixup\cite{mixup} is a data augmentation-based regularization that synthesizes new samples by linearly interpolating between pairs of samples and their labels in the input~\cite{mixup} or feature space~\cite{manifold_mixup,noisymix}. This approach increases the diversity of the training data, helps avoid overfitting, and encourages smoothing of the decision boundaries. However, it has also been pointed out that Mixup has the risk of excessive blurring of the boundaries between classes and propagation of noise, potentially degrading performance. As a follow-up to Mixup, Logit Mixing Training~\cite{logit_mix} introduced a regularization technique that enforces linearity not only in the input space but also in the logit space. In the context of OOD detection, Mixture Outlier Exposure (MixOE)\cite{MixOE} has been proposed as a representative method that applies Mixup to both ID and OOD samples while training, resulting in assigning high confidence to ID inputs and low confidence to OOD ones. However, the interpolated samples between ID and OOD data may become ambiguous and obscure the boundaries, leading to a decline in anomaly detection performance~\cite{mixup_boundary}.

In this study, we focus on the logit space, where decision boundaries between classes are more explicitly represented compared to the input space, in order to enhance the OOD detection performance via the Mixup strategy. Specifically, we propose Logit Mixture Outlier Exposure (Logit MixOE), a method built upon MixOE that performs Mixup in the logit space. Unlike MixOE, which interpolates in the input space, Logit MixOE linearly interpolates between ID and OOD representations in the logit space. This encourages smoother transitions between classes and strengthens detection performance, particularly against OOD samples that closely resemble ID data. Furthermore, inspired by the idea of Logit Mixing Training, we incorporate a regularization that enforces consistency between logits for OE: one obtained by directly mixing the logits of individual ID and OOD samples, and the other derived from feeding the sample interpolated in image space into the model. This consistency constraint helps suppress abrupt changes in model outputs near decision boundaries and promotes smoother and more discriminative boundaries for distinguishing OOD samples. 

In the experiments, we empirically evaluated both the baseline MixOE and the proposed Logit MixOE regarding classification for ID data and OOD detection. Experimental results show that while MixOE outperforms the proposed Logit MixOE under certain settings, further analysis reveals that enforcing our consistency regularization leads to improved detection performance. Moreover, we observed a trade-off between classification and detection performance depending on the mixing coefficient, which determines the interpolation ratio in both input and logit spaces. These findings suggest that further investigation into regularization strategies in the logit space for OOD detection is a promising direction for future work.
\section{Preliminaries}
\label{sec:preliminaries}

\subsection{Notation and Problem Formulation}
Let $\mathcal{D}_{\text{in}}$ denote a labeled training dataset consisting of pairs of a sample and its corresponding label independently drawn from the input space $\mathcal{X}$ and the label space $\mathcal{Y}$. This dataset is referred to as in-distribution~(ID) data. In contrast, a dataset containing samples drawn from a distribution that does not belong to the label space of the ID data is referred to as out-of-distribution~(OOD) data and is denoted by $\mathcal{D}_{\text{out}}$. In a standard $K$-class classification task, we are given the input data $\bm{x}\in\mathbb{R}^d$ and their corresponding one-hot encoded labels $\bm{y}$ from $\mathcal{D}_{\text{in}}$. The classification model $f:\mathbb{R}^d\rightarrow\mathbb{R}^K$ outputs a logit vector: 
\begin{align}
f(\bm{x})=[f_1(\bm{x}), f_2(\bm{x}), \dots, f_K(\bm{x})],
\end{align}
which is converted into a probability distribution via the Softmax function. The model is trained by minimizing the cross-entropy loss function $\ell_{\mathrm{ce}}$ between the label distribution and the predictive distribution:
\begin{align}
\ell_{\text{ce}}(f(\bm{x}), \bm{y})= -\sum_{i=1}^{K} y_i \log \frac{\exp(f_i(\bm{x}))}{\sum_{j=1}^{K}\exp(f_j(\bm{x}))}.
\end{align}
Typically, the model parameters are optimized by minimizing the empirical risk over a mini-batch $\mathcal{B}$, defined as: 
\begin{align}
\mathcal{L} = \frac{1}{|\mathcal{B}|}\sum_{(\bm{x}, \bm{y})\in\mathcal{B}}\ell_{\mathrm{ce}}\left(f(\bm{x}),\bm{y}\right).
\end{align}

In the OOD detection task, the goal is to identify whether a given input is from $\mathcal{D}_{\text{in}}$ or $\mathcal{D}_{\text{out}}$. To this end, an OOD score $S$ is calculated based on the classification model $f$ information. The detection of OOD sample is performed according to a pre-defined threshold $\tau$ as follows:
\begin{align}
S=\begin{cases}
\text{in}, & \text{if } S \geq \tau \\
\text{out}, & \text{if } S < \tau
\end{cases}.
\label{ood_def}
\end{align}

\subsection{Regularization for In-Distribution Data}
Firstly, we describe two regularization techniques proposed for in-distribution data $\mathcal{D}_{\text{in}}$: Mixup and Logit Mixing Training.

\overviewlogitmix

\subsubsection{Mixup}
Mixup is one of the regularization methods that trains the models on mixed samples generated by interpolating between pairs of samples in the input space. Specifically, given a mini-batch $\mathcal{B}$ sampled from in-distribution data $\mathcal{D}_{\text{in}}$, two samples $(\bm{x}_i,\bm{y}_i)$ and $(\bm{x}_j,\bm{y}_j)$ randomly selected, and a new mixed sample is generated via the following interpolation:
\begin{align}
\tilde{\bm{x}} &= \lambda_{\mathrm{input}} \bm{x}_i + \left( 1 - \lambda_{\mathrm{input}} \right)\bm{x}_j,\quad\lambda_{\mathrm{input}} \sim \mathrm{Beta}(\alpha, \alpha),
\end{align}
where $\lambda_{\mathrm{input}}\in[0,1]$ is the mixing coefficient, drawn from a Beta distribution parameterized by a hyperparameter $\alpha\in(0,\infty)$. The model is then trained to minimize the following mixup loss $\ell_{\mathrm{mix}}$, which linearly combines the cross-entropy losses of the two original labels:
\begin{align}
\ell_{\mathrm{mix}} = \lambda_{\mathrm{input}} \ell_{\text{ce}}(f(\tilde{\bm{x}}), \bm{y}_i) + \left( 1 - \lambda_{\mathrm{input}} \right)\ell_{\text{ce}}(f(\tilde{\bm{x}}), \bm{y}_j),
\end{align}
where $f(\tilde{\bm{x}})$ is the logit vector obtained by feeding the mixed sample $\tilde{\bm{x}}$ into the model.

\subsubsection{Logit Mixing Training}
Logit Mixing Training~\cite{logit_mix} is a regularization method that imposes a constraint to preserve linearity of the mixed input data in the logit space. Specifically, it introduces the following similarity loss:
\begin{align}
\ell_{\mathrm{sim}} = \left\|\left(\lambda_{\mathrm{logit}} f\left(\bm{x}_i\right)+(1-\lambda_{\mathrm{logit}}) f\left(\bm{x}_j\right)\right)-f\left(\tilde{\bm{x}}\right)\right\|_2,
\end{align}
where $\lambda_{\mathrm{logit}}$ is sampled from a Beta distribution parameterized by $\alpha\in(0,\infty)$, as in standard Mixup, to interpolate logits vectors in the logit space. This loss $\ell_{\mathrm{sim}}$ encourages the logit of the mixed input $\tilde{\bm{x}}$ to match the linear combination of the individual logits $f\left(\bm{x}_i\right)$ and $f\left(\bm{x}_j\right)$. This constraint enforces consistency between interpolation in the input space and the corresponding behavior in the logit space.

In the paper~\cite{logit_mix}, the model is trained using a combination of the mixup loss, the similarity loss, and the standard cross-entropy loss for classification:
\begin{align}
\ell_{\mathrm{cls}} = \ell_{\text{ce}}(f(\bm{x}_i), \bm{y}_i) + \ell_{\text{ce}}(f(\bm{x}_j), \bm{y}_j).
\end{align}

\subsection{Regularization for Out-of-Distribution Data}
Next, we describe two regularization methods designed for out-of-distribution data $\mathcal{D}_{\text{out}}$: Outlier Exposure and MixOE.

\subsubsection{Outlier Exposure}
Outlier Exposure (OE)~\cite{OE} is a regularization approach that improves the generalization performance and the robustness to OOD samples by intentionally exposing the model to auxiliary OOD data during training, in addition to ID samples. Let $\mathcal{D}_{\text{out}}^{\text{aux}}\subset \mathcal{D}_{\text{out}}$ denote the auxiliary OOD dataset used for training. Given an $(\bm{x}_{\mathrm{in}},\bm{y}_{\mathrm{in}})\sim \mathcal{D}_{\text{in}}$ and $\bm{x}_{\mathrm{out}}\sim \mathcal{D}_{\text{out}}^{\text{aux}}$, the OE encourage the model to assign low probabilities (confidence) to OOD inputs by minimizing following loss:
\begin{align}
\ell_{\mathrm{oe}} = \ell_{\text{ce}}(f(\bm{x}_{\mathrm{in}}), \bm{y}_{\mathrm{in}}) + \beta \ell_{\text{ce}}(f(\bm{x}_{\mathrm{out}}),  \mathcal{U}),
\end{align}
where $\beta$ is a weighting coefficient that balances the ID classification loss and the OOD regularization term, and $\mathcal{U}$ denotes a uniform distribution over all classes. This encourages the model to produce near-uniform predictions for OOD inputs, thereby making them distinguishable from ID ones.

\subsubsection{Mixture Outlier Exposure}
Mixture Outlier Exposure (MixOE) ~\cite{MixOE} is a method that extends OE by incorporating Mixup-based interpolation between ID and OOD data to further improve OOD detection performance. Specifically, MixOE performs interpolation between ID and OOD samples as follows:
\begin{align}
\tilde{\bm{x}}_{\mathrm{oe}}&=\lambda_{\mathrm{input}} \bm{x}_{\mathrm{in}}+(1-\lambda_{\mathrm{input}}) \bm{x}_{\mathrm{out}}, \\
\tilde{\bm{y}}_{\mathrm{oe}}&=\lambda_{\mathrm{input}} \bm{y}_{\mathrm{in}}+(1-\lambda_{\mathrm{input}}) \mathcal{U}.
\end{align}

The model is then trained using the following MixOE loss:
\begin{align}
\ell_{\mathrm{mixoe}} = \ell_{\text{ce}}(f(\bm{x}_{\mathrm{in}}), \bm{y}_{\mathrm{in}}) + \beta \ell_{\text{ce}}(f(\tilde{\bm{x}}_{\mathrm{oe}}),\tilde{\bm{y}}_{\mathrm{oe}}),
\label{eq:mixoe}
\end{align}
which encourages the model to make high-confidence predictions for ID data and low-confidence, i.e., near-uniform predictions for the interpolated samples between ID and OOD data.

\section{Proposed Method}
In this work, we aim to improve the OOD detection performance of MixOE by introducing regularization in the logit space. Specifically, our objectives are to strengthen the model's robustness to OOD data in MixOE, improve the detection performance of hard OOD data near the ID manifold, encourage smoother inter-class relations, and prevent overconfidence on OOD data. To achieve this, we propose two key techniques: Mixup in the logit space and a regularization term that enforces consistency between the mixed logits and those obtained from mixed input samples.

\subsection{Mixup in the Logit space}
Given ID data $(\bm{x}_{\mathrm{in}},\bm{y}_{\mathrm{in}})$ and OOD data $\bm{x}_{\mathrm{out}}$, we obtain their logit vectors $f(\bm{x}_{\mathrm{in}})$ and $f(\bm{x}_{\mathrm{out}})$ using a model $f$ pretrained on ID data. We then perform Mixup directly in the logit space as follows:
\begin{align}
\tilde{f}(\bm{x})&=\lambda_{\mathrm{logit}} f(\bm{x}_{\mathrm{in}})+(1-\lambda_{\mathrm{logit}}) f(\bm{x}_{\mathrm{out}}).
\end{align}

The model is then trained by minimizing a loss that combines the ID classification loss with the cross-entropy loss on the mixed logits:
\begin{align}
\ell_{\mathrm{logitmixoe}} = \ell_{\text{ce}}(f(\bm{x}_{\mathrm{in}}), \bm{y}_{\mathrm{in}}) + \beta \ell_{\text{ce}}(\tilde{f}(\bm{x}),\tilde{\bm{y}}_{\mathrm{oe}}).
\end{align}

\result

Unlike MixOE, which performs interpolation in the input space (Eq.~\eqref{eq:mixoe}), our approach mixes ID and OOD samples in the logit space. This is motivated by the fact that OOD samples trained with Outlier Exposure tend to produce nearly uniform softmax outputs, resulting in logits that are close to zero or constant across all classes. This property makes the distinction between ID and OOD samples more apparent in the logit space than in the input space. We refer to this model as Logit Mixture Outlier Exposure (Logit MixOE).

\subsection{Regularization on Logit Consistency}
Inspired by Logit Mixing Training, we also incorporate a regularization term that encourages consistency between the mixed logits in logit space and the logits obtained from feeding mixed inputs into the model for ID and OOD domains. This is defined as:
\begin{align}
\ell_{\mathrm{sim\_oe}} = \left\|\tilde{f}(\bm{x})-f\left(\tilde{\bm{x}}_{\mathrm{oe}}\right)\right\|_2.
\end{align}

The final training objective combines both terms:
\begin{align}
\ell_{\mathrm{total}} = \ell_{\mathrm{logitmixoe}} + \ell_{\mathrm{sim\_oe}}.
\end{align}

\figlogithist
\figlogitpca

\section{Experiment}
\label{sec:experiment}

\subsection{Experimental Setup}
\subsubsection{Evaluation Tasks and Metrics}
In order to validate the effectiveness of our regularization, we evaluate our method on two tasks: classification of in-distribution (ID) data and detection of out-of-distribution (OOD) data. Classification performance is measured using accuracy, while OOD detection performance is evaluated using AUROC and FPR95. A higher AUROC (closer to 1) indicates better performance, whereas a lower FPR95 (closer to 0) is preferred.

\subsubsection{Dataset}
Following the experimental setup in MixOE\cite{MixOE}, we used the North American Birds\cite{NABirdsD93:online}, FGVC-aircraft\cite{FGVC_aircraft}, and Butterfly\cite{ Butterfl33:online}. Hereafter, we refer to these datasets as \texttt{Bird}, \texttt{Aircraft}, and \texttt{Butterfly}, respectively. These datasets are publicly available fine-grained image recognition datasets. For each dataset, we adopt a hold-out class evaluation method~\cite{MixOE} to split the data into ID and OOD classes. ID samples are used for training, while the held-out classes are used as OOD data for evaluation. As the auxiliary OOD dataset $\mathcal{D}_\mathrm{out}^{aux}$ used during training, we employ WebVision 1.0\cite{webvision}, which contains approximately 2.4 million images collected from Google and Flickr and aligned with ImageNet classes. We apply filtering to \texttt{WebVision} to remove any samples belonging to \texttt{Bird}, \texttt{Bird}, \texttt{ Aircraft}, and \texttt{Butterfly} categories before using it in training.

\subsubsection{Training Details}
We performed pertaining and fine-tuning procedures using ResNet50~\cite{ResNet}. For pretraining, the model is trained for 90 epochs using SGD with a batch size of $32$, an initial learning rate of $0.001$, and a weight decay of $0.00001$. The learning rate is scheduled by cosine annealing. We apply standard data augmentations, including RandomHorizontalFlip and ColourJitter. For fine-tuning, the model is updated using ID data consisting of one of \texttt{Bird}, \texttt{Aircraft}, \texttt{Butterfly}, along with the filtered \texttt{WebVision} as the OOD dataset. Fine-tuning is performed for 10 epochs, using the same hyperparameters as in the pertaining phase.

\subsection{OOD Detection Performance}

Table~\ref{tb:result} shows the results of OOD detection and ID classification performance. In the following discussion, we focus on the OOD detection scores at $\alpha=1.0$ reported in the paper MixOE~\cite{MixOE}. From Table ~\ref{tb:result}, we observed that MixOE outperforms Logit MixOE on all datasets in terms of AUROC and FPR95. However, we also found that adding the logit consistency regularization term $\ell_\mathrm{sim\_oe}$ leads to improved detection performance in all settings except for MixOE with the \texttt{Butterfly} dataset. Notably, Logit MixOE w/ $\ell_\mathrm{sim\_oe}$ achieved the highest AUROC on both the $\texttt{Bird}$ and $\texttt{Aircraft}$ datasets, and the best FPR95 on the $\texttt{Aircraft}$ and $\texttt{Butterfly}$ datasets, outperforming all other comparison methods.

\subsection{Classification Performance}
From Table~\ref{tb:result}, we discuss the classification performance on ID data. Comparing MixOE and Logit MixOE, we found that the MixOE decreased by $2.13\%$ on \texttt{Bird} and $2.96\%$ \texttt{Aircraft}, while achieving an improvement of $0.52\%$ on \texttt{Butterfly}. Additionally, introducing the logit consistency regularization $\ell_\mathrm{sim\_oe}$ led to further improvements: $0.52\%$ gain for MixOE and $2.48\%$ for Logit MixOE on \texttt{Aircraft}.

\subsection{Logit Visualisation}
To better understand the effectiveness of the proposed method, we analyze the distribution of logits. 

\subsubsection{Histogram of Logits}
We first visualize the histograms of the L2 norm of logit for both ID and OOD data using \texttt{Bird} dataset as the ID data, as shown in Figure~\ref{fig:logithist}. The figures show the results for the pretrained model, MixOE, Logit MixOE, MixOE w/ $\ell_\mathrm{sim\_oe}$, and Logit MixOE w/ $\ell_\mathrm{sim\_oe}$. In the pretrained model, since ID and OOD data are not explicitly separated during training, their distributions largely overlap. In contrast, after fine-tuning with MixOE, we observed a clearer separation between the two distributions. Although Logit MixOE is less separated than MixOE, introducing $\ell_\mathrm{sim\_oe}$ improves the separation between ID and OOD logits.

\figlogitlogitmixoe
\figlogitlogitmixoelsim

\subsubsection{PCA Visualization of Logits}
We further visualize the logits of ID and OOD data using PCA in Figure~\ref{fig:logitpca}. From the figure, it can be seen that the pre-trained models show significant variations in the OOD data features, while both MixOE and LogitMixOE tend to produce more compact representations for OOD samples. Moreover, the addition of $\ell_\mathrm{sim\_oe}$ reduces the overall logit scale and causes OOD samples to cluster around the center of the ID data. In other words, $\ell_\mathrm{sim\_oe}$ encourages that OE's properties, the logit of the OOD data have lower magnitudes than ID data, particularly pushing the OOD logits closer to zero. Consequently, the model may fail to preserve inter-class structure among ID classes, leading to a trade-off: classification accuracy decreases, while OOD detection performance improves.

\subsubsection{Sample-wise Logit Responses}
Finally, we visualise responses on a per-sample logit. Figures~\ref{fig:logit_logitmixoe} and \ref{fig:logit_logitmixoelsim} show the logits for one selected ID data, OOD data, input-space mixed data, and logit-space mixed data for Logit MixOE with and without $\ell_\mathrm{sim\_oe}$. From these figures, it can be seen that the logit of the OOD data tends to have lower magnitudes than ID data, highlighting the effect of the OE term. Additionally, Logit MixOE exhibits a reduction in logit magnitude when $\ell_\mathrm{sim\_oe}$ is applied, particularly pushing the OOD logits closer to zero.

\figbeta

\subsection{Effect of the Mixup Hyperparameters $\alpha$}
Figure~\ref{fig:beta} shows the Beta distributions corresponding to different values of the hyperparameter $\alpha$. As shown in the figure, when $\alpha$ is large, the sampled mixing coefficient $\lambda$ tends to be close to 0.5, resulting in mixed images and logits that are intermediate between ID data and OOD data. Conversely, when $\alpha$ is small, $\lambda$ tends to be near 0 or 1, making the mixed data to closely resemble either the ID or OOD samples.

Although MixOE also adjusts $\alpha$, identifying the optimal value of $\alpha$ for OOD detection remains an open question. Based on Table~\ref{tb:result}, we observed that adding $\ell_\mathrm{sim\_oe}$ improves OOD performance when larger $\alpha$ values are used. We attribute this to the model's ability to learn from intermediate samples between ID and OOD, which helps smooth the inter-class boundaries and enhances robustness to OOD inputs.

Furthermore, we observed a trade-off between classification and detection performance depending on the value of $\alpha$. Specifically, increasing $\alpha$ improves classification accuracy at the expense of OOD detection performance, while decreasing $\alpha$ enhances OOD detection but may degrade classification. Therefore, the $\alpha$ value is important and should be tuned based on the target application.

\section{CONCLUSIONS}
\label{sec:conclusion}
In this study, we proposed a combination of a novel OOD detection framework and regularization that regularizes the model’s output logits to improve the detection performance, generalization, and robustness of existing OOD detection models. Especially, we observed that introducing the logit consistency regularization term into our proposed Logit MixOE effectively pushes OOD logits closer to zero, thereby enhancing OOD detection.

Additionally, through experiments varying the Mixup parameter $\alpha$, we found a trade-off between classification and detection performance. This indicates that the value of $\alpha$ should be tuned according to the target application and the performance aspect to be prioritized. While adding $\ell_\mathrm{sim\_oe}$ improves OOD detection, it can also degrade classification performance. This highlights a key direction for future work: introducing constraints or adaptive mechanisms on $\ell_\mathrm{sim\_oe}$ to suppress only the logits of OOD samples while preserving the class-discriminative structure of ID data.

{\small
\bibliographystyle{ieee}
\bibliography{main}
}

\end{document}